\documentclass[ba]{imsart}
\pubyear{2021}
\volume{TBA}
\issue{TBA}
\firstpage{1}
\lastpage{1}

\usepackage{amsthm}
\usepackage{amsmath}
\usepackage{natbib}
\usepackage[colorlinks,citecolor=blue,urlcolor=blue,filecolor=blue,backref=page]{hyperref}
\usepackage{cleveref}
\usepackage{graphicx}
\usepackage{amsfonts}
\startlocaldefs
\numberwithin{equation}{section}
\theoremstyle{plain}

\endlocaldefs
\newcommand{\widesim}[2][1.5]{
  \mathrel{\overset{#2}{\scalebox{#1}[1]{$\sim$}}}
}
\usepackage{algorithm}
\usepackage[noend]{algpseudocode}
\usepackage{xr}
\usepackage{float}
\usepackage[caption = false]{subfig}

\begin{document}

\begin{frontmatter}
\title{Bayesian Feature Selection in Joint Quantile Time Series Analysis}
\runtitle{Bayesian Feature Selection}

\begin{aug}
\author{\fnms{Ning} \snm{Ning}\thanksref{addr1,t1}\ead[label=e1]{patning@tamu.edu}}

\runauthor{Ning Ning}

\address[addr1]{Department of Statistics, Texas A\&M University, College Station, TX. 
    \printead{e1} 
}

\thankstext{t1}{The research of Ning Ning was partially supported by the Seed Fund Grant Award at Texas A\&M University. The author would like to thank three anonymous reviewers and the Associate Editor for their very constructive comments and efforts on this work, which greatly improved the quality of this paper.}

\end{aug}

\begin{abstract}
Quantile feature selection over correlated multivariate time series data has always been a methodological challenge and is an open problem. In this paper, we propose a general Bayesian dimension reduction methodology for feature selection in high-dimensional joint quantile time series analysis, under the name of the quantile feature selection time series (QFSTS) model. The QFSTS model is a general structural time series model, where each component yields an additive contribution to the time series modeling with direct interpretations. Its flexibility is compound in the sense that users can add/deduct components for each time series and each time series can have its own specific valued components of different sizes. Feature selection is conducted in the quantile regression component, where each time series has its own pool of contemporaneous external predictors allowing nowcasting. Bayesian methodology in extending feature selection to the quantile time series research area is developed using multivariate asymmetric Laplace distribution, spike-and-slab prior setup, the Metropolis-Hastings algorithm, and the Bayesian model averaging technique, all implemented consistently in the Bayesian paradigm. The QFSTS model requires small datasets to train and converges fast. Extensive examinations confirmed that the QFSTS model has superior performance in feature selection, parameter estimation, and forecast.
\end{abstract}



\end{frontmatter}

\section{Introduction}
\label{Introduction}
We firstly give the background and motivation in Section \ref{sec:Background_and_motivation} and then state our contributions in Section \ref{sec:Our_contributions}, followed with the organization of the paper in Section \ref{sec:organization}.

\subsection{Background and motivation}
\label{sec:Background_and_motivation}
In the era of ``big data", electronic devices are now involved in many social activities, and can capture, store, manipulate, and analyze vast caches of such data. Conventional statistical and econometric techniques become increasingly inadequate to deal with ``big data" problems, such as the curse of dimensionality, large storage requirement, and
high computational cost (see, e.g. \cite{ning2021scalable}). Feature selection, also known as variable selection, attribute selection, or variable subset selection, is a machine learning technique for dimensionality reduction (see, e.g. \cite{lamnisos2009transdimensional}). It aims to choose a small subset of the
relevant features (variables, predictors) for use in model construction, from the original ones by removing irrelevant, redundant, or noisy features. It usually leads to better learning performance in terms of higher learning accuracy, lower computational cost, and better model interpretability. It has also been proven to be an effective and efficient way of shortening training times,
improving data's compatibility with a learning model class, and
encoding inherent symmetries present in the input space (see, e.g. \cite{griffin2021adaptive}).

Time series forecasting as one of the most applied data science techniques have been used extensively in finance, supply chain management, production and inventory planning, etc. However, as the time dimension adds additional information, time series problems are more difficult to handle compared to other prediction tasks (\cite{kalli2014time}).
 \cite{scott2014predicting, scott2015bayesian} developed the Bayesian Structural Time Series (BSTS) model, which can be used for time series forecasting, nowcasting, inferring causal relationships, etc. \cite{jammalamadaka2018multivariate} proposed the multivariate Bayesian structural time series (MBSTS) model, which extended the BSTS model to multivariate target time series with various components. The BSTS/MBSTS model has wide applications (see, e.g., \cite{Jammalamadaka2019Predicting,qiu2020multivariate} and the references therein). 

One can think of the future value of a univariate time series as a random variable whose distribution is the forecast distribution. Analogously, the future values of multivariate time series can be thought of as correlated random variables with the joint forecast distribution. The usual time series forecast is often to predict the mean or the median of the (joint) forecast distribution, which can be categorized as a point forecast. The (joint) quantile forecast is appealing in many economic applications (\cite{ley2007jointness}), such as value at risk in the finance industry in order to develop a strategy for trading and/or hedging purposes. Recently, \cite{griffin2022bayesian} proposed a Bayesian quantile time series model for asset returns which beautifully handled formal Bayesian inference on quantiles. It is an open question so far on how to incorporate the feature selection technique in joint quantile time series analysis, which is our goal of this paper.



\subsection{Our contributions}
\label{sec:Our_contributions}
In this paper, we propose a general Bayesian methodology for feature selection in joint quantile analysis with multivariate correlated time series data, under the name of quantile feature selection time series (QFSTS) model. Our contributions are four-fold:

\begin{itemize}
\item \textbf{General model structure.}	First, the QFSTS model is a 
structural time series model, which further belongs to state space models. All state components are assembled independently, and each component yields an additive contribution to the time series modeling with direct interpretations. The flexibility of the model allows users to include different  components for each target series, such as one time series has a trend component and a regression component while another correlated time series only has a regression component. Given that our main goal is to illustrate the feature selection methodology in joint quantile time series analysis, we focus on a trend component, a seasonal component,  a regression component, and an error term (equation \eqref{eq:st}) without loss of generality.

Second, the QFSTS model allows nowcasting and each time series can have its own pool of contemporaneous external predictors. Nowcasting is to forecast a current value instead of a future value (\cite{banbura2010nowcasting}). Two scenarios for using nowcasting are the following: First, many economic time series are reported infrequently such as on a monthly or quarterly basis, despite being theoretically observable on finer time scales; Second, time series are also frequently revised after they are first reported, as new information becomes available. An effective nowcasting model considers both the past behavior of the series being modeled and the values of more easily observed contemporaneous signals. All predictors in the regression component of the QFSTS model can be contemporaneous with a known lag by shifting the corresponding predictors in time. The QFSTS model allows each time series to have its own pool of predictors, for example one time series has $50$ predictors and another has $20$ different predictors. Feature selection is conducted among each times series' pool of predictors while still considering the correlations among different times series.

\item \textbf{Methodology.} First, the QFSTS model is the first on introducing the multivariate asymmetric Laplace (MAL) distribution to time series analysis. The idea of Bayesian quantile regression employing a likelihood function that is based on the asymmetric Laplace (AL) distribution, was introduced in the classical work by \cite{yu2001bayesian}. The use of the
AL distribution is proved as a very natural and effective way of modeling Bayesian quantile regression, irrespective of the original distribution of the data (\cite{chen2013bayesian}). In Section \ref{sec:Quantile_Regression_components}, we rigorously establish the explicit linkage between the MAL distribution and joint quantile regression, by setting the specific forms of parameters of the MAL distributed error term.

Second, the QFSTS model achieves feature selection in joint quantile time series analysis. It is very challenging to conduct feature selection on correlated time series where each has its own predictors, while when it comes to selecting features in quantile regression, it is much more challenging in that one has to take quantiles into consideration instead of the mean. Further, the Bayesian quantile regression coefficients depend on the quantile of interest (\cite{yu2001bayesian}). That is, for different quantiles, the coefficient of any predictor varies.
The QFSTS model uses the Gibbs sampler for quantile feature selection by means of the classical spike-and-slab prior setup (\cite{george1997approaches,madigan1994model}), and the Metropolis-Hastings algorithm. Bayesian variable selection can be performed using mixture priors with a spike and a slab component for the effects subject to selection, where the spike prior concentrates its mass at values close to zero allowing shrinkage of small effects to zero, and the slab prior has its mass spread over a wide range of plausible values for the regression coefficients. As the spike is concentrated at zero, variable selection is based on the probability of assigning the corresponding regression effect to the slab component. These posterior inclusion probabilities can be determined by MCMC sampling.

Third, the QFSTS model introduces the Bayesian model averaging technique (\cite{hoeting1999bayesian}) in joint quantile time series analysis. 
Bayesian model averaging provides a coherent mechanism to handle model uncertainty, by averaging the predicted quantile values from all the models generated in model training (\cite{fernandez2001benchmark, steel2020model}). 
In this way, we commit neither to any particular set of predictors which helps avoid an arbitrary selection, nor to point estimates of their coefficients which prevents overfitting.

\item \textbf{Excellent performance.} 
Under the challenging situation that each time series has its own pool of predictors made of both continuous and categorical covariates, the QFSTS model provides very accurate feature selection and parameter estimation results, thoroughly examined on simulated data, with different quantile values, different size of datasets, and different correlation values among multivariate time series.
Extensive analyses confirmed that the QFSTS model outperforms the ``auto.arima" function in ten steps of one-step-ahead quantile forecast consistently. The ``auto.arima" function automatically fits the autoregressive integrated moving average with regression (ARIMAX) model that is the most classical and popular time series model.

\item \textbf{Wide applicability.} First, because of the feature selection function embedded in the QFSTS model, it is applicable in proper applications that desire higher learning accuracy, lower computational cost, better model interpretability, and other benefits generated by feature selection.  
Second, because of the Bayesian paradigm embedded in the QFSTS model, it is applicable in proper applications that require Bayesian inference and learning. Third, because of the QFSTS model's general model structure, it is applicable in more applications besides the finance applications demonstrated in this paper. 
At last, the QFSTS model requires small datasets to train and converges fast.
	
\end{itemize}

\subsection{Organization of the paper}
\label{sec:organization}
The rest of the paper proceeds as follows:
In Section \ref{sec:model}, we introduce the QFSTS model by explaining its model structure and then illustrating the linkage between the specific design of the QFSTS model with multivariate quantile prediction.
In Section \ref{sec:Methodology}, we explain the methodology of the QFSTS model, by firstly writing the system in the distribution equivalence matrix form, and then providing the prior setup and posterior formulas. In Section \ref{sec:Algorithms}, we provide the model training algorithm and the joint quantile time series forecast algorithm. In Section \ref{sec:Model_Performance_with_Simulated_Data}, we demonstrate the model performance with simulated data generated by three-dimensional target time series, and fully examine the QFSTS model's ability in selecting the correct variables, accuracy in parameter estimation, and strength in forecast, with different quantiles and time series correlations.
In Section \ref{sec:conclusion}, we conclude and remark. In the Supplementary Material, we conduct further numerical analysis with simulated data and demonstrate the model performance with empirical data. 
Throughout the paper, the symbol $\tau$, with or without subscripts, will represent the quantile level.

\section{The QFSTS model}
\label{sec:model}
In this section, we introduce the QFSTS model, by firstly explaining its model structure in Section \ref{sec:time_series_components}, and then illustrating the linkage between the specific design of the QFSTS model with multivariate quantile prediction in Section \ref{sec:Quantile_Regression_components}.

\subsection{Model Structure}
\label{sec:time_series_components}
The QFSTS model is a 
structural time series model, which is
constructed by components with direct interpretations, as follows:
\begin{equation} \label{eq:st}
\widetilde{y}_t=\widetilde{\mu}_t+\widetilde{\iota}_t+\widetilde{\xi}_t+\widetilde{\epsilon}_t,
\end{equation}
where $t$ stands for a specific time point in $\{1,\cdots,n\}$, the $m$-dimensional vector $\widetilde{y}_t$ represents target time series where $m>1$, the $m$-dimensional vector $\widetilde{\mu}_t$ represents the linear trend component, the $m$-dimensional vector $\widetilde{\iota}_t$ represents the seasonal component, the $m$-dimensional vector $\widetilde{\xi}_t$ represents the regression component,
and the $m$-dimensional vector $\widetilde{\epsilon}_t$ represents the observation error term. Since structural time series models belong to state space models, the QFSTS model
then belongs to state space models. In the following, we illustrate each time series components:
\begin{itemize}
	\item The linear trend component $\widetilde{\mu}_t=[\mu_{t}^{(1)},\cdots,\mu_{t}^{(m)}]^T$ of the QFSTS model is a generalization of the local linear trend model, in the following form:
	\begin{align} \label{eq:trend}
	\mu_{t+1}^{(i)}=\mu_t^{(i)}+\delta_t^{(i)}+u_t^{(i)}, \quad\quad[u_{t}^{(1)},\cdots,u_{t}^{(m)}]^T\widesim{\text{iid}} \  N_m(0,\Sigma_{\widetilde{\mu}}),\\
	\label{eq:slope}
	\delta_{t+1}^{(i)}=D_i+\lambda_i(\delta_t^{(i)}-D_i)+v_t^{(i)},\quad\quad[v_{t}^{(1)},\cdots,v_{t}^{(m)}]^T\widesim{\text{iid}} \  N_m(0,\Sigma_{\widetilde{\delta}}).
	\end{align}
	The QFSTS model allows each target series to have its own specific linear trend component.
	Here, for the $i$-th target series where $i\in \{1,\cdots,m\}$, $\mu_t^{(i)}$ models the current ``level" of the trend; $\delta_t^{(i)}$ models the current ``slope" of the trend which is the expected increase in $\mu_t^{(i)}$ between time point $t$ and time point $t+1$; the parameter $D_i$ models the ``mean reversion" effect, i.e., a dynamic can be very unstable in the short run but stable in the long run; the parameter $\lambda_i\in [0,1]$ is the learning rate at which the local trend is updated.
	
	\item The seasonal component $\widetilde{\iota}_t=[\iota_{t}^{(1)},\cdots,\iota_{t}^{(m)}]^T$ of the QFSTS model in the following form is one frequently used model:
	\begin{equation} \label{eq:seasonal}
		\iota_{t+1}^{(i)}=-\sum_{k=0}^{S_i-2}{\iota}^{(i)}_{t-k}+w^{(i)}_t,  \quad\quad [w^{(1)}_t,\cdots,w^{(m)}_t]^T\widesim{\text{iid}}   \ N_m(0,\Sigma_\iota),
	\end{equation}
	where $S_i$ represents the number of seasons for $y^{(i)}$. The QFSTS model allows each target series to have its own specific seasonal component.
	For instance, we might include a seasonal component with $S_i=7$ to capture day-of-the-week effect for target series $y^{(i)}$, and $S_j=30$ indicating day-of-the-month effect for another target series $y^{(j)}$ when modeling daily data. 
	
	\item The regression component $\widetilde{\xi}_t=[\xi_{t}^{(1)},\cdots,\xi_{t}^{(m)}]^T$ of the QFSTS model  is written as follows:
	\begin{equation} \label{eq:regression}
	\xi^{(i)}_t=\beta_i^Tx^{(i)}_t.
	\end{equation}
	The QFSTS model allows each target series to have its own specific regression component. Here for $i\in \{1,\cdots,m\}$,  $x_t^{(i)}=[x_{t1}^{(i)},\cdots,x_{tk_i}^{(i)}]^T$ is the pool of all available $k_i$ predictors at time $t$ for the $i$-th target series, for example $k_i=30$ for the $i$-th target series and $k_j=20$ for the $j$-th target series where $j\neq i$;  $\beta_i=[\beta_{i1},\cdots,\beta_{ik_i}]^T$ represents corresponding static regression coefficients for the $i$-th target series. 
	
	\item The vector of observation error terms $\widetilde{\epsilon}_t=[\epsilon_{t}^{(1)},\cdots,\epsilon_{t}^{(m)}]^T$ follows the multivariate asymmetric Laplace (MAL) distribution 
	\begin{equation} \label{eq:error}
	\widetilde{\epsilon}_t\widesim{\text{iid}} \mathcal{AL}_m(\phi_\epsilon,\Sigma_\epsilon),
	\end{equation}
	where $\phi_\epsilon$ is a $m$-dimensional vector of means for each variable, and $\Sigma_\epsilon$ is a $m\times m$-dimensional non-negative definite symmetric matrix of variance-covariance.
	When $\phi_\epsilon=0$, the distribution $\mathcal{AL}_m(0,\Sigma_\epsilon)$ is the symmetric multivariate Laplace distribution. We refer interested readers to Section $6.2$ of \cite{kotz2012laplace} for further explanations of the MAL distribution.
\end{itemize}

\subsection{The MAL Distribution and Joint Quantile Regression}
\label{sec:Quantile_Regression_components}
In this section, we establish the linkage between the MAL distribution and joint quantile regression, by setting the specific forms of $\phi_\epsilon$ and $\Sigma_\epsilon$ in the $\mathcal{AL}_m(\phi_\epsilon,\Sigma_\epsilon)$ distribution, which is given in equation \eqref{eqn:link}. 

Firstly, we subtract the trend component and the seasonal component from the multivariate target time series and denote
$$\widetilde{z}_t=[z_{t}^{(1)},\cdots,z_{t}^{(m)}]^T=\widetilde{y}_t-\widetilde{\mu}_t-\widetilde{\iota}_t.$$
Then equation \eqref{eq:st} becomes
\begin{align}
\label{eqn:quantile_regression_exp1}
\widetilde{z}_t=\widetilde{\xi}_t+\widetilde{\epsilon}_t,   \quad \quad \widetilde{\epsilon}_t\widesim{\text{iid}} \mathcal{AL}_m(\phi_\epsilon,\Sigma_\epsilon). 
\end{align}
By Theorem $6.3.1$ in Section $6.3$ of \cite{kotz2012laplace}, $\widetilde{\epsilon}_t$ can be represented as follows, in the distribution equivalence sense,
\begin{equation} 
\label{eqn:quantile_regression_exp2}
\widetilde{\epsilon}_t=\phi_\epsilon W+W^{1/2}\widetilde{e}_t, \quad\quad W\widesim{\text{iid}} \operatorname{Exp}(1), \quad\quad \widetilde{e}_t=[e_{t}^{(1)},\cdots,e_{t}^{(m)}]^T\widesim{\text{iid}}  N_m(0,\Sigma_\epsilon),
\end{equation}
where $W$ and $\widetilde{e}_t$ are independent,
therefore we can rewrite \eqref{eqn:quantile_regression_exp1} as 
\begin{equation}\label{eqn:quantile_regression_component}
\widetilde{z}_t=\widetilde{\xi}_t+\phi_\epsilon W+W^{1/2}\widetilde{e}_t,
\end{equation}
in the distribution equivalence sense.

The specific forms of $\phi_\epsilon$ and $\Sigma_\epsilon$ in the $\mathcal{AL}_m(\phi_\epsilon,\Sigma_\epsilon)$ distribution which link 
the MAL distribution with joint quantile regression, is based on the corresponding theory in the univariate case. Now, we summarize some of the findings in \cite{yu2001bayesian}. The standard linear regression model is given by
$$y_t^u=(\boldsymbol{x}_t)^T \boldsymbol{\beta}+\epsilon_t^u,$$
where the superscript ``$u$"  indicates the univariate case and bold symbols represent vectors.
Here, $\boldsymbol{x}_t$ is the vector of regressors; $\boldsymbol{\beta}$ is the vector of corresponding coefficients; the error term $\epsilon_t^u$ has zero mean and finite constant variance, but it is not necessary to specify its distribution as it can take any form.   
Suppose that the $p$-th ($0<p<1$) quantile of the error term $\epsilon_t^u$ is the value $q_p$, such that $\mathbb{P}(\epsilon_t^u<q_p)=p$. The $p$-th conditional quantile of $y_t^u$ given $\boldsymbol{x}_t$ is then simply
\begin{equation}
\label{eqn:uni-quantile_reg}
q_p(y_t^u\mid \boldsymbol{x}_t)= (\boldsymbol{x}_t)^T \boldsymbol{\beta}_p,
\end{equation}
where $\boldsymbol{\beta}_p$ is a vector of coefficients dependent on $p$. 

The $p$-th regression quantile is defined as the solution $\widehat{\boldsymbol{\beta}}_p$ to the quantile regression minimization problem
$$\min_{\boldsymbol{\beta}}\sum_t \rho_p\bigg(y_t^u-(\boldsymbol{x}_t)^T \boldsymbol{\beta}\bigg),$$
where $\rho_p(\cdot)$ is a loss function that has robust properties (see, \cite{huber2004robust}) given by
\begin{align}
\label{eqn:quantile_loss_func}
\rho_p(u)=\frac{|u|+(2p-1)u}{2}.
\end{align}
\cite{yu2001bayesian} showed that the minimization of the above loss function is exactly equivalent to the maximization
of a likelihood function formed by combining independently distributed asymmetric Laplace (AL) densities, and the univariate AL distribution provides a direct connection between the quantile regression minimization problem and the maximum likelihood estimation. 

A random variable $U$ is said to follow the univariate AL distribution $\mathcal{AL}(\theta_{\text{loc}},\theta_{\text{sca}},p)$ 
if its probability density function is given by
$$f_p(u)=\frac{p(1-p)}{\theta_{\text{sca}}}\exp\left\{-\rho_p\left(\frac{u-\theta_{\text{loc}}}{\theta_{\text{sca}}}\right)\right\},$$
where $\rho_p(\cdot)$ is given in \eqref{eqn:quantile_loss_func}, $\theta_{\text{loc}}$ is the location parameter, and $\theta_{\text{sca}}$ is the scale parameter. The mean and the variance of $U$ are given by
$$\mathbb{E}(U)=\theta_{\text{loc}}+\theta_{\text{sca}}\frac{1-2p}{p(1-p)},\quad\quad \operatorname{Var}(U)=\theta_{\text{sca}}^2\frac{1-2p+2p^2}{p^2(1-p)^2}.$$ 
Now we get back to equation \eqref{eqn:quantile_regression_component} and investigate its univariate case of the $i$-th target series for $i\in \{1,2,\cdots, m\}$:
\begin{equation}
z_t^{(i)}=\xi_t^{(i)}+\phi_\epsilon^{(i)} W+W^{1/2}e_{t}^{(i)},   \quad    \quad  \quad 
e_{t}^{(i)}\widesim{\text{iid}}  \mathcal {N}(0,(\Sigma_\epsilon)_{ii}),
\end{equation}
Suppose we are interested in the $\tau_i$-th $(0<\tau_i<1)$ quantile and then we need 
$z_t^{(i)}$ to satisfy the univariate AL distribution $\mathcal{AL}(\xi_t^{(i)},\phi_i,\tau_i)$.

In order to obtain
$$\mathbb{E}(z_t^{(i)})=\xi_t^{(i)}+\phi_i\frac{1-2\tau_i}{\tau_i(1-\tau_i)},$$
it suffices to set 
\begin{align}
\label{eqn:vector_condition}
\phi_\epsilon^{(i)}=\phi_i\frac{1-2\tau_i}{\tau_i(1-\tau_i)}
\end{align}
since 
$$\mathbb{E}\left(\phi_\epsilon^{(i)} W\right)=\phi_\epsilon^{(i)}\quad\text{and}\quad \mathbb{E}\left(W^{1/2}e_{t}^{(i)}\right)=0.$$
Furthermore, in order to obtain
$$\operatorname{Var}(z_t^{(i)})=(\phi_i)^2\frac{1-2\tau_i+2\tau_i^2}{\tau_i(1-\tau_i)},$$
it suffices to set 
\begin{align}
\label{eqn:matrix_condition}
(\Sigma_\epsilon)_{ii}=(\phi_i)^2\frac{2}{\tau_i(1-\tau_i)},
\end{align}
since then
\begin{align*}
\operatorname{Var}(z_t^{(i)})=&\operatorname{Var}(\phi_\epsilon^{(i)} W+W^{1/2}e_{t}^{(i)})\\
=&\left(\phi_i\frac{1-2\tau_i}{\tau_i(1-\tau_i)}\right)^2+(\phi_i)^2\frac{2}{\tau_i(1-\tau_i)}\\
=&(\phi_i)^2\frac{1-2\tau_i+2\tau_i^2}{\tau_i(1-\tau_i)}.
\end{align*} 
At last, in order to meet the univariate requirements \eqref{eqn:vector_condition}
and \eqref{eqn:matrix_condition}, we can take
\begin{align}
\label{eqn:link}
\phi_\epsilon=\Phi \widetilde{\phi}_{\tau} \quad \text{and}\quad \Sigma_\epsilon=\Phi\Sigma_{\tau}\Phi=\Phi(\Psi_{\tau}\Sigma_{\text{corr}}\Psi_{\tau})\Phi,
\end{align}
where $\widetilde{\phi}_{\tau}=\left[\frac{1-2\tau_1}{\tau_1(1-\tau_1)},\frac{1-2\tau_2}{\tau_2(1-\tau_2)},\cdots,\frac{1-2\tau_m}{\tau_m(1-\tau_m)}\right]^T$, $\Sigma_{\text{corr}}$ is the correlation matrix, 
\begin{equation*} 
\begin{split}
\Phi=\begin{bmatrix}
\phi_1 & 0  & \dots  & 0 \\
0 & \phi_2  & \dots  & 0 \\
\vdots & \vdots  & \ddots & \vdots \\
0 & 0 & \dots  & \phi_{m}
\end{bmatrix},
\Psi_{\tau}=\begin{bmatrix}
\sqrt{\frac{2}{\tau_1(1-\tau_1)}} & 0  & \dots  & 0 \\
0 & \sqrt{\frac{2}{\tau_2(1-\tau_2)}}  & \dots  & 0 \\
\vdots & \vdots  & \ddots & \vdots \\
0 & 0  & \dots  & \sqrt{\frac{2}{\tau_m(1-\tau_m)}}
\end{bmatrix}.
\end{split}
\end{equation*}

\section{Methodology}
\label{sec:Methodology}
In this section, we explain the methodology of the QFSTS model.  
In Section \ref{sec:Equivalence_Matrix_Form}, we write the system in the distribution equivalence matrix form to facilitate derivations following.
In Section \ref{sec:prior}, we provide the prior setups including 
the spike-and-slab  for variable selection in this Bayesian paradigm.
In Section \ref{sec:posterior}, we derive all necessary conditional probabilities, in order to implement the classical Gibbs sampler in model training.

\subsection{The Distribution Equivalence Matrix Form}
\label{sec:Equivalence_Matrix_Form}

Recall that by equation \eqref{eqn:quantile_regression_component} we have
$$
\widetilde{z}_t=\widetilde{\xi}_t+\phi_\epsilon W+W^{1/2}\widetilde{e}_t.
$$
Here, $\widetilde{z}_t=[z_{t}^{(1)},\cdots,z_{t}^{(m)}]^T$ is the $m$-dimensional vector generated by subtracting the trend component and the seasonal component from the multivariate target time series; 
$\widetilde{\xi}_t=[\xi_{t}^{(1)},\cdots,\xi_{t}^{(m)}]^T$  is the $m$-dimensional  regression component such that
$\xi^{(i)}_t=\beta_i^Tx^{(i)}_t$, where
$\beta_i=[\beta_{i,1},\cdots,\beta_{i,k_i}]^T$ is the collection of regression coefficients for the $i$-th response variable, and $x_t^{(i)}=\left[x_{t1}^{(i)},\dots,x_{tk_i}^{(i)}\right]^T$ is the pool of all available $k_i$ predictors at time $t$ for the $i$-th target series; $\phi_\epsilon$ is the $m$-dimensional vector of means for the $m$-dimensional error term.

Now, we define the $n\times m$ matrix $Z$ as
$$Z=[\widetilde{z}_1,\dots,\widetilde{z}_n]^T=\begin{bmatrix}
\begin{bmatrix}
z_{1}^{(1)}\\\vdots\\z_{1}^{(m)}
\end{bmatrix}, \cdots, \begin{bmatrix}
z_{n}^{(1)}\\\vdots\\z_{n}^{(m)}
\end{bmatrix}
\end{bmatrix}^T
=
\begin{bmatrix}
\begin{bmatrix}
z_{1}^{(1)}&\cdots &z_{1}^{(m)}
\end{bmatrix}\\ \cdots \\\begin{bmatrix}
z_{n}^{(1)}&\cdots &z_{n}^{(m)}
\end{bmatrix}
\end{bmatrix},$$ 
and then define the $mn$-dimensional vector $\widetilde{Z}$ as 
$$\widetilde{Z}=\operatorname{vec}(Z)=
\begin{bmatrix}
\begin{bmatrix}
z_{1}^{(1)}&\cdots &z_{n}^{(1)}
\end{bmatrix}, \cdots, \begin{bmatrix}
z_{1}^{(m)}&\cdots &z_{n}^{(m)}
\end{bmatrix}
\end{bmatrix}^T.$$  
Analogously, we define the $mn$-dimensional vector $\widetilde{E}$ as
$$\widetilde{E}=\operatorname{vec}(E)\quad \text{where} \quad E=[\widetilde{\epsilon}_1,\dots,\widetilde{\epsilon}_n]^T,$$ and define the $mn$-dimensional vector $\widetilde{\Phi}_\epsilon$ as
$$\widetilde{\Phi}_\epsilon=\operatorname{vec}(\Phi_\epsilon)\quad \text{where} \quad\Phi_\epsilon=[\phi_\epsilon,\dots,\phi_\epsilon]^T.$$
Define $\beta$ as the $K$-dimensional $(K=\sum_{i=1}^{m}k_i)$ vector of regression coefficients  $$\beta=\left[\beta_{1}^T,\cdots,\beta_{m}^T\right]^T\quad \text{where} \quad \beta_i=[\beta_{i,1},\cdots,\beta_{i,k_i}]^T.$$
Define $X_i$ as the $n \times k_i$ matrix representing all observations of $k_i$ possible 
predictors for the $i$-th response variable
\begin{equation} \label{eq:Xj}
X_i=\left[(x_1^{(i)})^T,\cdots,(x_n^{(i)})^T\right]^T=\begin{bmatrix}
x_{1,1}^{(i)} & x_{1,2}^{(i)} & x_{1,3}^{(i)} & \dots  & x_{1,k_i}^{(i)} \\
x_{2,1}^{(i)} & x_{2,2}^{(i)} & x_{2,i,3}^{(i)} & \dots  & x_{2,k_i}^{(i)} \\
\vdots & \vdots & \vdots & \ddots & \vdots \\
x_{n,1}^{(i)} & x_{n,2}^{(i)} & x_{n,3}^{(i)} & \dots  & x_{n,k_i}^{(i)}
\end{bmatrix},
\end{equation}
and further define $X$ as the $mn\times K$-dimensional predictor matrix 
\begin{equation} \label{eq:X}
X=\begin{bmatrix}
X_1 & 0 & 0 & \dots  & 0 \\
0 & X_2 & 0 & \dots  & 0 \\
\vdots & \vdots & \vdots & \ddots & \vdots \\
0 & 0 & 0 & \dots  & X_{m}
\end{bmatrix}.
\end{equation}
Then we have the following expression, in the distribution equivalence sense,
\begin{equation}\label{eqn:quantile_regression_matrix}
\widetilde{Z}=X\beta+\widetilde{\Phi}_\epsilon W+W^{1/2}\widetilde{E}.
\end{equation}

\subsection{Prior Distributions}
\label{sec:prior}
The spike-and-slab prior setup is a Bayesian variable selection technique. 
To specify spike prior, a vector of $K=\sum_{i=1}^{m}k_i$ indicator variables 
$$\gamma=[\gamma_{1,1},\cdots,\gamma_{1,k_1},\;\gamma_{2,1},\cdots,\gamma_{2,k_2},\;\cdots,\;\gamma_{m,1},\cdots,\gamma_{m,k_m}]$$ is introduced according to the rule:
\begin{equation*} 
\gamma_{i,k}= 
\begin{cases}
1 & \text{if}\ \beta_{i,k}\ne 0,\\
0        & \text{otherwise}.
\end{cases}
\end{equation*}
Regressor indicators are assumed to be independent Bernoulli variables:
\begin{equation} \label{eq:ind} 
p(\gamma|W)=\prod_{i=1}^{m}\prod_{k=1}^{k_i}\pi_{i,k}^{\gamma_{i,k}}(1-\pi_{i,k})^{1-\gamma_{i,k}},\ \ \ \  \ \ 0 \leq \pi_{i,k}=p(\gamma_{i,k})\leq 1,
\end{equation}
where $\pi_{i,k}$ is the prior inclusion probability. 
Equation \eqref{eq:ind} is often simplified by setting $\pi_{i,k}=\pi_i$, if prior information of specific predictors on response variables is not available. One could further simplify by setting $\pi_i=q_i/k_i$, where $k_i$ is the total number of candidate predictors for the $i$-th target series and $q_i$ nonzero expected predictors given by researchers. When there is sufficient prior information, assigning subjectively determined values to $\pi_{i,k}$ might provide more robust results. One could also force certain variables to be excluded or included
by setting $\pi_{ij}$ as $0$ or $1$. By default and in all the experimental examinations following, we set $\pi_{ik}=0.5$ for all $i\in \{1,\cdots,m\}$ and all $k\in \{1,\cdots,k_i\}$.


We use
a simple conventional prior specification which makes $\beta$ and $\Sigma_{\epsilon}$ conditionally independent 
\begin{equation} \label{eqn:conditionally_independent}
p(\beta,\Sigma_{\tau}|\gamma)=p(\beta|\gamma)\times p(\Sigma_{\tau}|\gamma),
\end{equation}
where
\begin{equation} \label{eqn:slab}
\beta|\gamma \widesim{\text{iid}} N_K(b_\gamma,A_\gamma^{-1}),  \quad\quad \quad
\Sigma_{\tau}|\gamma \widesim{\text{iid}} IW_m(v_0,V_0).
\end{equation}
Equation \eqref{eqn:slab} is
the slab prior because, conditional on $\gamma$, one can choose the prior parameters to make it only very weakly informative and close to flat.
$N_K(b_\gamma,A_\gamma^{-1})$ stands for the $K$-dimensional multivariate normal distribution, $b_\gamma$ is the vector of prior means, and  $A_\gamma$ is the full-model prior information matrix. One can set
$A_\gamma=\kappa X^T_\gamma X_\gamma/n$ where $\kappa$ is the number of observations worth of weight on the prior mean vector $b_\gamma$. 
$IW_m(v_0, V_0)$ stands for the $m$-dimensional inverse Wishart (IW) distribution, where $v_0$ is the number of degrees of freedom and $V_0$ is a $m\times m$ scale matrix. One can ask analysts for an expected $R^2$, and a number of observations worth of weight $v_0$ which must be greater than the dimension of $\widetilde{y}_t$ plus one, and set $$V_0=(v_0-m-1)(1-R^2)\Sigma_y,$$ where $\Sigma_y$ is the variance-covariance matrix for multiple target time series. For simplicity, we set $b_\gamma=0$, $\kappa=0.01$, $R^2=0.8$, and $v_0=5$  in all the experimental examinations following. We acknowledge that the IW prior setup may not necessarily conform to the specific form presented in Equation \eqref{eqn:link}, which is one example demonstrating the connection to the univariate case. However, considering the difficulty in estimating the correlation matrix for multivariate time series, we opt for the IW prior as a simpler alternative. Consequently, it is important to note that our modeling approach may not be optimal, and there is potential for further improvements to be made over our methodology.

Since we are going to use the Metropolis-Hastings Algorithm (see Chapter $6.3.1$ of \cite{robert2010introducing}) to learn the distribution of $\Phi$, we allow the prior distribution of its elements to be any distribution that is proportional to $1$. The prior distributions of variance-covariance matrices in the trend component and the seasonal component are set as the inverse Wishart distribution
\begin{equation} \label{eq:prior_state_variance}
\Sigma_{\alpha} \widesim{\text{iid}} IW_m(\nu_\alpha,V_\alpha),\quad\quad \alpha\in \{\widetilde{\mu},\widetilde{\delta},\widetilde{\iota} \}. 
\end{equation}
For simplicity, we set 
$\nu_{\alpha}=V_{\alpha}=0.01$  in all the experimental examinations following. 

\subsection{Posterior Conditional Distributions}
\label{sec:posterior}
In order to implement the classical Gibbs sampler in this multivariate setting,  we derive all necessary conditional probabilities of $\widetilde{Z}$, $\beta$, $\Phi$, $\Sigma_{\tau}$, $\gamma$, and $W$.
The full likelihood function under model assumptions is given by
\begin{align*} 
p(\widetilde{Z},\beta,\Phi,\Sigma_{\tau},\gamma,W) =p(\widetilde{Z}|\beta,\Phi,\Sigma_{\tau},\gamma,W)\times p(\beta,\Sigma_{\tau}|\gamma)\times p(\gamma)\times p(W)\times p(\Phi).
\end{align*}
Then, by equations \eqref{eqn:quantile_regression_matrix} -- \eqref{eqn:slab} and the setup of $p(\Phi)$, we have that
\begin{align} 
\label{eq:liklihood}
&p(\widetilde{Z},\beta,\Phi,\Sigma_{\tau},\gamma,W)\nonumber\\
\propto & p(\widetilde{Z}|\beta,\Phi,\Sigma_{\tau},\gamma,W)\times p(\beta|\gamma,W)\times p(\Sigma_{\tau}|\gamma,W)\times p(\gamma)\times p(W)\nonumber\\
\propto &|W\Phi\Sigma_{\tau}\Phi|^{-n/2} \exp \left(-\frac{1}{2W}(\widetilde{Z}-X_{\gamma}\beta_{\gamma}-\widetilde{\Phi}_\epsilon W)^T((\Phi\Sigma_{\tau}\Phi)^{-1}\otimes I_n)(\widetilde{Z}-X_{\gamma}\beta_{\gamma}-\widetilde{\Phi}_\epsilon W)\right)\nonumber\\
& \times |A_\gamma|^{1/2}\exp\left(-\frac{1}{2}(\beta_\gamma-b_\gamma)^T A_\gamma(\beta_\gamma-b_\gamma)\right)  |\Sigma_{\tau}|^{-(v_0+m+1)/2}\exp\left(-\frac{1}{2}\operatorname{tr}(V_0\Sigma_{\tau}^{-1})\right)\nonumber\\
& \times p(\gamma)\times p(W),
\end{align}
where $|\cdot|$ stands for the determinant of a matrix, $\otimes$ is the Kronecker product, and $\operatorname{tr}(\cdot)$ represents the trace of a matrix. 

\subsubsection{Posterior Conditional Distribution of $\beta$}
To facilitate derivation, we firstly transform 
$$\widetilde{Z}=X\beta+\widetilde{\Phi}_\epsilon W+W^{1/2}\widetilde{E},$$
where 
$$\widetilde{E}=\operatorname{vec}(E)=\operatorname{vec}( [\widetilde{\epsilon}_1,\dots,\widetilde{\epsilon}_n]^T), \quad\quad \widetilde{e}_t\widesim{\text{iid}}  N_m(0,\Sigma_\epsilon=\Phi\Sigma_{\tau}\Phi),$$
to a system with uncorrelated errors using the Cholesky decomposition of $\Sigma_{\tau}$, 
\begin{equation} \label{eq:Cholesky}
\Sigma_{\tau}=U^TU,\quad \quad\text{i.e.}\; (U^{-1})^T\Sigma_{\tau} U^{-1}=I.
\end{equation}
Thus we have the transformed system with uncorrelated errors:
\begin{equation} \label{eq:standard}
\widehat{Z}=\widehat{X}\beta+\widehat{\Phi}_{\epsilon}W+W^{1/2}\widehat{E},
\end{equation}
where
\begin{equation} \label{eq:standard2}
\begin{gathered}\widehat{Z}=(((U\Phi)^{-1})^T\otimes I_n)\widetilde{Z},  \quad\quad\widehat{X}=(((U\Phi)^{-1})^T\otimes I_n)X,\\
\widehat{\Phi}_{\epsilon}=(((U\Phi)^{-1})^T\otimes I_n)\widetilde{\Phi}_{\epsilon},\quad\quad
\widehat{E}=(((U\Phi)^{-1})^T\otimes I_n)\widetilde{E}.
\end{gathered}
\end{equation}
For the following term in the first exponential in \eqref{eq:liklihood}, we have
\begin{equation}
\label{eqn:decorrelation} 
\begin{split}
&(\widetilde{Z}-X_{\gamma}\beta_{\gamma}-\widetilde{\Phi}_\epsilon W)^T((\Phi\Sigma_{\tau}\Phi)^{-1}\otimes I_n)(\widetilde{Z}-X_{\gamma}\beta_{\gamma}-\widetilde{\Phi}_\epsilon W)\\
=&(\widetilde{Z}-X_{\gamma}\beta_{\gamma}-\widetilde{\Phi}_\epsilon W)^T(([U\Phi] ^TU\Phi)^{-1}\otimes I_n)(\widetilde{Z}-X_{\gamma}\beta_{\gamma}-\widetilde{\Phi}_\epsilon W)\\
=&(\widetilde{Z}-X_{\gamma}\beta_{\gamma}-\widetilde{\Phi}_\epsilon W)^T((U\Phi)^{-1}\otimes I_n)\times (((U\Phi)^{-1})^T\otimes I_n)(\widetilde{Z}-X_{\gamma}\beta_{\gamma}-\widetilde{\Phi}_\epsilon W)\\
=&(\widehat{Z}-\widehat{X}_{\gamma} \beta_{\gamma}-\widehat{\Phi}_{\epsilon}W)^T(\widehat{Z}-\widehat{X}_{\gamma} \beta_{\gamma}-\widehat{\Phi}_{\epsilon}W).
\end{split}
\end{equation}

The full conditional distribution of $\beta$ can be expressed as:
\begin{equation*} 
\begin{split}
p(\beta|\widehat{Z},\Phi,\Sigma_{\epsilon},\gamma,W)
\propto &\exp\left(-\frac{1}{2}W^{-1}(\widehat{Z}-\widehat{X}_{\gamma} \beta_{\gamma}-\widehat{\Phi}_{\epsilon}W)^T(\widehat{Z}-\widehat{X}_{\gamma} \beta_{\gamma}-\widehat{\Phi}_{\epsilon}W)\right)\\
&\times \exp\left(-\frac{1}{2}(\beta_\gamma-b_\gamma)^T A_\gamma(\beta_\gamma-b_\gamma)\right).
\end{split}
\end{equation*}
Terms in the above exponential can be written as
\begin{equation}
\label{eq:exp_term_beta}
\begin{split}
&W^{-1}(\widehat{Z}-\widehat{X}_{\gamma} \beta_{\gamma}-\widehat{\Phi}_{\epsilon}W)^T(\widehat{Z}-\widehat{X}_{\gamma} \beta_{\gamma}-\widehat{\Phi}_{\epsilon}W)+(\beta_\gamma-b_\gamma)^T A_\gamma(\beta_\gamma-b_\gamma)\\
=&\beta_\gamma^T(W^{-1}\widehat{X}_\gamma^T\widehat{X}_\gamma+A_\gamma)\beta_\gamma-\beta_\gamma^T(W^{-1}\widehat{X}^T\widehat{Z}-\widehat{X}^T\widehat{\Phi}_{\epsilon} +A_\gamma b_\gamma)\\
&-(W^{-1}\widehat{X}_\gamma^T\widehat{Z}-\widehat{X}_\gamma^T\widehat{\Phi}_{\epsilon}+A_\gamma b_\gamma)^T\beta_\gamma+W^{-1}(\widehat{Z}-\widehat{\Phi}_{\epsilon}W)^T(\widehat{Z}-\widehat{\Phi}_{\epsilon}W)+b_\gamma^T A_\gamma b_\gamma\\
=&(\beta_\gamma-\overline{\beta}_{\gamma})^T(W^{-1}\widehat{X}_\gamma^T\widehat{X}_\gamma+A_\gamma)(\beta_\gamma-\overline{\beta}_{\gamma})+W^{-1}(\widehat{Z}-\widehat{\Phi}_{\epsilon}W)^T(\widehat{Z}-\widehat{\Phi}_{\epsilon}W)\\
&+b_\gamma^T A_\gamma b_\gamma-(\overline{\beta}_{\gamma})^T(W^{-1}\widehat{X}_\gamma^T\widehat{X}_\gamma+A_\gamma)\overline{\beta}_{\gamma},
\end{split}
\end{equation}
where 
$$\overline{\beta}_{\gamma}=(W^{-1}\widehat{X}_\gamma^T\widehat{X}_\gamma+A_\gamma)^{-1}(W^{-1}\widehat{X}_\gamma^T\widehat{Z}-\widehat{X}_\gamma^T\widehat{\Phi}_{\epsilon}+A_\gamma b_\gamma).$$
Therefore, $\beta$ is still conditionally multivariate normal distributed
\begin{equation} \label{eq:pbeta}
\begin{gathered}
\beta|\widehat{Z},\Phi,\Sigma_{\tau},\gamma,W \widesim{\text{iid}} N_K(\overline{\beta}_{\gamma},(W^{-1}\widehat{X}_\gamma^T\widehat{X}_\gamma+A_\gamma)^{-1}).
\end{gathered}
\end{equation}

\subsubsection{Posterior Conditional Distribution of $\Sigma_{\tau}$}

Recalling that $X_i$ is the $n \times k_i$-dimensional matrix given in equation \eqref{eq:Xj}, define the $n\times K$-dimensional ($K=\sum_{i=1}^{m}k_i$) matrix $X^{\ast}_\gamma$ as 
$$X^{\ast}_\gamma=[X_1,X_2,\dots,X_m]_{\gamma}.$$
Define the $K\times m$-dimensional matrix $B_\gamma$ as
\begin{equation*} 
B_\gamma=\begin{bmatrix}
\beta_1 & 0 & 0 & \dots  & 0 \\
0 & \beta_2 & 0 & \dots  & 0 \\
\vdots & \vdots & \vdots & \ddots & \vdots \\
0 & 0 & 0 & \dots  & \beta_{m}
\end{bmatrix}_{\gamma}, \quad\quad  \beta_i=\begin{bmatrix}\beta_{i,1}\\ \vdots\\\beta_{i,k_i}\end{bmatrix},
\end{equation*}
where $\beta_i$ is the $k_i$-dimensional vector containing the collection of regression coefficients for the $i$-th response series. 
For the reason that trace is invariant under cyclic permutations, from equation \eqref{eq:liklihood}, we know that
\begin{align*}
&(\widetilde{Z}-X_\gamma \beta_\gamma-\widetilde{\Phi}_\epsilon W)^T((\Phi\Sigma_{\tau}\Phi)\otimes I_n)^{-1}(\widetilde{Z}-X_\gamma \beta_\gamma-\widetilde{\Phi}_\epsilon W)\\ 
&=\operatorname{vec}\bigg(Z-X^{\ast}_\gamma B_\gamma-\Phi_\epsilon W\bigg)^T((\Phi\Sigma_{\tau}\Phi)^{-1}\otimes I_n)\operatorname{vec}\bigg(Z-X^{\ast}_\gamma B_\gamma-\Phi_\epsilon W\bigg)\\
&= \operatorname{tr}\bigg((Z-X^{\ast}_\gamma B_\gamma-\Phi_\epsilon W)^T 
(Z-X^{\ast}_\gamma B_\gamma-\Phi_\epsilon W)\Phi^{-1}\Sigma_{\tau}^{-1}\Phi^{-1}\bigg)\\
&= \operatorname{tr}\bigg(
\bigg[(Z-X^{\ast}_\gamma B_\gamma-\Phi_\epsilon W)\Phi^{-1}\bigg]\Sigma_{\tau}^{-1} \bigg[(Z-X^{\ast}_\gamma B_\gamma-\Phi_\epsilon W)\Phi^{-1}\bigg]^T \bigg)\\
&= \operatorname{tr}\bigg(\bigg[(Z-X^{\ast}_\gamma B_\gamma-\Phi_\epsilon W)\Phi^{-1}\bigg]^T
\bigg[(Z-X^{\ast}_\gamma B_\gamma-\Phi_\epsilon W)\Phi^{-1}\bigg]\Sigma_{\tau}^{-1}  \bigg),
\end{align*}
and then we have
\begin{align*} 
&p(\Sigma_{\tau}|\widetilde{Z},\Phi,\beta,\gamma,W)\\
\propto & |\Sigma_{\tau}|^{-(n+v_0+m+1)/2} \exp\bigg(-\frac{1}{2}\operatorname{tr}\bigg(\frac{1}{W}\bigg[(Z-X^{\ast}_\gamma B_\gamma-\Phi_\epsilon W)\Phi^{-1}\bigg]^T\\
&\hspace*{3.7cm}\times\bigg[(Z-X^{\ast}_\gamma B_\gamma-\Phi_\epsilon W)\Phi^{-1}\bigg]\Sigma_{\tau}^{-1}+V_0 \Sigma_{\tau}^{-1}\bigg)\bigg).
\end{align*}
That is, the posterior conditional distribution of $\Sigma_{\tau}$ is in the invert Wishart form
\begin{align} \label{eq:psigma}
&\Sigma_{\tau}|\widetilde{Z},\beta,\Phi,\gamma,W\\
& \widesim{\text{iid}} IW_m\left(v_0+n,\frac{1}{W}\bigg[(Z-X^{\ast}_\gamma B_\gamma-\Phi_\epsilon W)\Phi^{-1}\bigg]^T\bigg[(Z-X^{\ast}_\gamma B_\gamma-\Phi_\epsilon W)\Phi^{-1}\bigg]+V_0\right).\nonumber
\end{align}

\subsubsection{Posterior Conditional Distribution of $\Phi$}


Recall that by the Cholesky decomposition in \eqref{eq:Cholesky} we have that 
$\Sigma_{\tau}=U^TU$.
Further recall that by \eqref{eqn:link} we have that the $m$-dimensional vector
$\phi_\epsilon=\Phi \widetilde{\phi}_{\tau}$ where $\Phi$ is a $m\times m$-dimensional diagonal matrix and $\widetilde{\phi}_{\tau}$ is a $m$-dimensional vector, and then we can write the $n\times m$-dimensional matrix $\Phi_\epsilon$ as
$$\Phi_\epsilon=[\phi_\epsilon,\dots,\phi_\epsilon]^T=\widetilde{\Phi}_\tau \Phi, \quad\quad\text{where}\quad \widetilde{\Phi}_\tau=[\widetilde{\phi}_{\tau},\cdots, \widetilde{\phi}_{\tau}]^T. $$
Then by \eqref{eq:liklihood} we have
\begin{align} 
\label{eqn:phi_posterior}
&p(\Phi|\widetilde{Z},\beta,\Sigma_{\tau},\gamma,W)\nonumber\\
\propto&|\Phi|^{-n}\exp\left(-\frac{1}{2}
\operatorname{tr}\bigg(\frac{1}{W}\bigg[(Z-X^{\ast}_\gamma B_\gamma-\Phi_\epsilon W)\Phi^{-1}\bigg]^T\right.\nonumber\\
&\hspace*{4.2cm}\times\left.
\bigg[(Z-X^{\ast}_\gamma B_\gamma-\Phi_\epsilon W)\Phi^{-1}\bigg]\Sigma_{\tau}^{-1}  \bigg)\right)\nonumber\\
\propto&|\Phi|^{-n}\exp\left(-\frac{1}{2}
\operatorname{tr}\bigg(\frac{1}{W}\bigg[(Z-X^{\ast}_\gamma B_\gamma-\Phi_\epsilon W)\Phi^{-1}\bigg]^T\right.\nonumber\\
&\hspace*{4.2cm}\times\left.
\bigg[(Z-X^{\ast}_\gamma B_\gamma-\Phi_\epsilon W)\Phi^{-1}\bigg]U^{-1}(U^{-1})^T  \bigg)\right)\nonumber\\
\propto&|\Phi|^{-n}\exp\left(-\frac{1}{2}
\operatorname{tr}\bigg(\frac{1}{W}\bigg[(Z-X^{\ast}_\gamma B_\gamma-\Phi_\epsilon W)\Phi^{-1}U^{-1}\bigg]^T\right.\nonumber\\
&\hspace*{4.2cm}\times\left.
\bigg[(Z-X^{\ast}_\gamma B_\gamma-\Phi_\epsilon W)\Phi^{-1}U^{-1}\bigg] \bigg)\right)\nonumber\\
\propto&|\Phi|^{-n}\exp\left(-\frac{1}{2}
\operatorname{tr}\bigg(\frac{1}{W}\bigg[(Z-X^{\ast}_\gamma B_\gamma-\widetilde{\Phi}_\tau \Phi W)\Phi^{-1}U^{-1}\bigg]^T\right.\nonumber\\
&\hspace*{4.2cm}\times\left.\bigg[(Z-X^{\ast}_\gamma B_\gamma-\widetilde{\Phi}_\tau \Phi W)\Phi^{-1}U^{-1}\bigg] \bigg)\right)\nonumber\\
\propto&|\Phi|^{-n}\exp\left(-\frac{1}{2}
\operatorname{tr}\bigg(\frac{1}{W}\bigg[(Z-X^{\ast}_\gamma B_\gamma)\Phi^{-1}U^{-1}-\widetilde{\Phi}_\tau U^{-1}W\bigg]^T\right.\\
&\hspace*{4.2cm}\times\left.
\bigg[(Z-X^{\ast}_\gamma B_\gamma)\Phi^{-1}U^{-1}-\widetilde{\Phi}_\tau U^{-1}W\bigg] \bigg)\right)\nonumber
\end{align}
%

\subsubsection{Posterior Conditional Distribution of $\gamma$}
By equations \eqref{eq:liklihood} and \eqref{eq:exp_term_beta}, we know that
\begin{align*} 
\label{eq:liklihood2}
&p(\widetilde{Z},\beta,\Phi,\Sigma_{\tau},\gamma,W)\\
\propto &\exp\left(-\frac{1}{2}\left[(\beta-\overline{\beta}_{\gamma})^T(W^{-1}\widehat{X}_\gamma^T\widehat{X}_\gamma+A_\gamma)(\beta-\overline{\beta}_{\gamma})+W^{-1}(\widehat{Z}-\widehat{\Phi}_{\epsilon}W)^T(\widehat{Z}-\widehat{\Phi}_{\epsilon}W)\right]\right)\\
& \times |A_\gamma|^{1/2}\exp\left(-\frac{1}{2}\left[\operatorname{tr}(V_0\Sigma_{\epsilon}^{-1})+b_\gamma^T A_\gamma b_\gamma-(\overline{\beta}_{\gamma})^T(W^{-1}\widehat{X}_\gamma^T\widehat{X}_\gamma+A_\gamma)\overline{\beta}_{\gamma}\right]\right)p(\gamma)\\
& \times|W\Phi\Sigma_{\tau}\Phi|^{-n/2}|\Sigma_{\tau}|^{-(v_0+m+1)/2},
\end{align*}
where 
$$\overline{\beta}_{\gamma}=(W^{-1}\widehat{X}_\gamma^T\widehat{X}_\gamma+A_\gamma)^{-1}(W^{-1}\widehat{X}_\gamma^T\widehat{Z}-\widehat{X}_\gamma^T\widehat{\Phi}_{\epsilon}+A_\gamma b_\gamma).$$
Furthermore, by the fact that 
$$\beta|\widehat{Z},\Phi,\Sigma_{\tau},\gamma,W \widesim{\text{iid}} N_K(\overline{\beta}_{\gamma},(W^{-1}\widehat{X}_\gamma^T\widehat{X}_\gamma+A_\gamma)^{-1}),$$ 
we have
\begin{align*} 
\int_{-\infty}^{\infty} \exp\left(-\frac{1}{2}(\beta-\overline{\beta}_{\gamma})^T(W^{-1}\widehat{X}_\gamma^T\widehat{X}_\gamma+A_\gamma)(\beta-\overline{\beta}_{\gamma})\right) d\beta\propto |W^{-1}\widehat{X}_\gamma^T\widehat{X}_\gamma+A_\gamma|^{-1/2},
\end{align*}
and then
\begin{equation*} 
\begin{split}
&p(\widetilde{Z},\Phi,\Sigma_{\tau},\gamma,W)\\
=&\int_{-\infty}^{\infty}
p(\widetilde{Z},\beta,\Phi,\Sigma_{\tau},\gamma,W)d\beta\\
\propto &\exp\left(-\frac{1}{2}\left[W^{-1}(\widehat{Z}-\widehat{\Phi}_{\epsilon}W)^T(\widehat{Z}-\widehat{\Phi}_{\epsilon}W)-\Xi_{\gamma}^T(W^{-1}\widehat{X}_\gamma^T\widehat{X}_\gamma+A_\gamma)^{-1}\Xi_{\gamma}\right]\right)\\
& \times | A_\gamma|^{1/2}\exp\left(-\frac{1}{2}\left[\operatorname{tr}(V_0\Sigma_{\epsilon}^{-1})+b_\gamma^T A_\gamma b_\gamma\right]\right)
|W^{-1}\widehat{X}_\gamma^T\widehat{X}_\gamma+A_\gamma|^{-1/2}
p(\gamma)\\
& \times|W\Phi\Sigma_{\tau}\Phi|^{-n/2}|\Sigma_{\tau}|^{-(v_0+m+1)/2} p(W),
\end{split}
\end{equation*}
where 
$$\Xi_{\gamma}=(W^{-1}\widehat{X}_\gamma^T\widehat{Z}-\widehat{X}_\gamma^T\widehat{\Phi}_{\epsilon}+A_\gamma b_\gamma).$$
Therefore, the posterior conditional distribution of $\gamma$ is given by
\begin{equation} 
\label{eq:pgamma}
\begin{split}
p(\gamma | \widetilde{Z},\Phi,\Sigma_{\tau},W)
\propto &\exp\left(-\frac{1}{2}\left[b_\gamma^T A_\gamma b_\gamma-\Xi_{\gamma}^T(W^{-1}\widehat{X}_\gamma^T\widehat{X}_\gamma+A_\gamma)^{-1}\Xi_{\gamma}\right]\right)\\
& \times |A_\gamma|^{1/2}  |W^{-1}\widehat{X}_\gamma^T\widehat{X}_\gamma+A_\gamma|^{-1/2}p(\gamma).
\end{split}
\end{equation}

\subsubsection{Posterior Conditional Distribution of $W$}
Recall that the generalized inverse Gaussian distribution (GIG) is a three-parameter family of continuous probability distributions with probability density function (see page $1$ of \cite{jorgensen2012statistical})
$$ f(x)={\frac {(a/b)^{p/2}}{2K_{p}({\sqrt {ab}})}}x^{(p-1)}e^{-(ax+b/x)/2},\qquad x>0,$$
where $K_p$ is a modified Bessel function of the second kind, $a > 0$, $b > 0$, and $p$ is a real parameter. 
By equations \eqref{eq:liklihood} and \eqref{eqn:decorrelation}, we have that
\begin{equation*} 
\begin{split}
&p(W|\widetilde{Z},\beta,\widetilde{\Phi},\Sigma_{\epsilon},\gamma)\\
&\propto |W|^{-n/2} \exp \left(-\frac{1}{2W}(\widehat{Z}-\widehat{X}_{\gamma} \beta_{\gamma}-\widehat{\Phi}_{\epsilon}W)^T(\widehat{Z}-\widehat{X}_{\gamma} \beta_{\gamma}-\widehat{\Phi}_{\epsilon}W)-W\right),
\end{split}
\end{equation*}
based on which,
\begin{equation} 
\label{eq:pW}
W|\widetilde{Z},\beta,\Sigma_{\epsilon},\gamma \widesim{\text{iid}} \operatorname{GIG}(a,b,p),
\end{equation}
\begin{equation*} 
a=2+ \widehat{\Phi}_{\epsilon}^T\widehat{\Phi}_{\epsilon}, \quad\quad  b= (\widehat{Z}-\widehat{X}_{\gamma} \beta_{\gamma})^T(\widehat{Z}-\widehat{X}_{\gamma} \beta_{\gamma}), \quad\quad  p=1-n/2. 
\end{equation*}

\subsubsection{Posterior Conditional Distribution of $\Sigma_\alpha$}
Next we need to derive conditional posterior distribution of $\Sigma_\alpha$ where $\alpha\in \{\widetilde{\mu},\widetilde{\delta},\widetilde{\iota}\}$ in the trend component and the seasonal component. Similarly, as the posterior conditional distribution of $\Sigma_{\tau}$ in the invert Wishart form in equation \eqref{eq:psigma}, the posterior distribution of $\Sigma_\alpha$ is conditionally inverse Wishart distributed
\begin{equation} \label{eq:30}
\Sigma_\alpha|\widetilde{Y},\alpha,W \widesim{\text{iid}} IW_m\left(\nu_\alpha+n,V_\alpha+\frac{1}{W}AA^T\right),\quad \quad\quad\alpha\in \{\widetilde{\mu},\widetilde{\delta},\widetilde{\iota}\}, 
\end{equation}
where $A$ is the matrix of a collection of residues of each time series component.

\section{Algorithms}
\label{sec:Algorithms}
Gibbs sampling is a Markov chain Monte Carlo (MCMC) algorithm for obtaining a sequence of observations, which are approximated from a specified multivariate probability distribution.
MCMC methods are to construct a Markov chain that has the desired distribution as its equilibrium distribution. One can draw samples of the desired distribution by discarding the initial MCMC steps as ``burn-in'', since the quality of  samples is an
increasing function of the number of steps. 
In Algorithm \ref{algo:slide_generator}, the posterior distributions of the model are simulated by Gibbs sampling approach, in the way that looping through the $7$ steps yields a sequence of draws $\theta=(\alpha,\Sigma_\alpha,\beta,\Phi,\Sigma_{\tau},\gamma,W)$ where $\alpha\in \{\widetilde{\mu},\widetilde{\delta},\widetilde{\iota}\}$, from a Markov chain with the stationary probability distribution $p(\theta|Y)$ which is the posterior distribution of $\theta$ given $Y$.

\begin{algorithm}[!ht]
	\caption{Model Training}
	\label{algo:slide_generator}
	
	\begin{algorithmic}[1]
		\bigskip
		\Statex \textbf{Time series state components}
		\State Draw the latent state $\alpha$ from $p(\alpha|\widetilde{Y},\Sigma_\alpha,\beta,\Phi,\Sigma_{\tau},\gamma,W)$ where $\alpha\in \{\widetilde{\mu},\widetilde{\delta},\widetilde{\iota}\}$, using the posterior simulation algorithm from \cite{durbin2002simple}.
		
		\State Draw time series state component parameters $\Sigma_\alpha$ from
		$\Sigma_\alpha \widesim{\text{iid}} p(\Sigma_\alpha|\widetilde{Y},\alpha,W)$ based on the inverse Wishart distribution in equation \eqref{eq:30}.\bigskip
		
		\Statex \textbf{Quantile regression component }
		\State Loop over $i$ in an random order, draw each  $\gamma_i|\gamma_{-i},\widetilde{Z},\Phi,\Sigma_{\tau},W$, namely simulating $\gamma \widesim{\text{iid}} p(\gamma | \widetilde{Z},\Phi,\Sigma_{\tau},W)$  in equation \eqref{eq:pgamma}, using the stochastic search variable selection (SSVS) algorithm from \cite{george1997approaches}.
		\State Draw $\beta$ from $ \beta \widesim{\text{iid}} p(\beta|\widehat{Z},\Phi,\Sigma_{\tau},\gamma,W)$ based on the multivariate normal distribution in equation \eqref{eq:pbeta}.\bigskip
		
		\Statex \textbf{Error term}	
		\State Draw $\Sigma_{\tau}$ from $\Sigma_{\tau} \widesim{\text{iid}} p(\Sigma_{\tau}|\widetilde{Z},\beta,\Phi,\gamma,W)$ based on the inverse Wishart distribution in equation \eqref{eq:psigma}.
		\State Draw $\Phi$ based on  $p(\Phi|\widetilde{Z},\beta,\Sigma_{\tau},\gamma,W)$ in equation \eqref{eqn:phi_posterior} using the Metropolis-Hastings Algorithm.
		\State Draw $W$ from $W \widesim{\text{iid}} p(W|\widetilde{Z},\beta,\Sigma_{\epsilon},\gamma)$ based on the generalized inverse Gaussian distribution in equation \eqref{eq:pW}.
		\bigskip
	\end{algorithmic}
\end{algorithm}

\begin{algorithm}[!ht]
	\caption{Joint Quantile Predictions}
	\label{algo:forecast}
	\begin{algorithmic}[1]
		\bigskip
		\State Draw the next trend component $\alpha_{t+1}=(\widetilde{\mu}_{t+1},\widetilde{\delta}_{t+1},\widetilde{\iota}_{t+1})$, given current trend component $\alpha_{t}=(\widetilde{\mu}_{t},\widetilde{\delta}_{t},\widetilde{\iota}_{t})$ and variance-covariance parameters $\Sigma_\alpha=(\Sigma_{\widetilde{\mu}},\Sigma_{\widetilde{\delta}},\Sigma_{\widetilde{\iota}})$, by equations \eqref{eq:trend} and \eqref{eq:slope}.
		\State Based on indicator variable $\gamma$, compute the regression component $\widetilde{\xi}_{t+1}$ given the information about predictors at time $t+1$, by equation \eqref{eq:regression}.
		\State Draw a random error $\widetilde{\epsilon}_{t+1}$ in the multivariate asymmetric Laplace distribution by equation \eqref{eq:error}, whose mean and variance are generated by expressions given in equation \eqref{eqn:link}.
		\State Sum up $\widetilde{\mu}_{t+1}$, $\widetilde{\iota}_{t+1}$, $\widetilde{\xi}_{t+1}$, and $\widetilde{\epsilon}_{t+1}$ to generate predictions, by equation \eqref{eq:st}. 
		\State Sum up all the generated predictions and divide by the total number of effective MCMC iterations to generate the joint quantile predictions.
		\bigskip 
	\end{algorithmic}
\end{algorithm}

Given draws of model parameters and latent states from their posterior distributions, we can draw samples from the posterior predictive distribution 
\begin{equation*} 
p(\widehat{Y}|Y)=\int p(\widehat{Y}|\theta)p(\theta|Y)d\theta,
\end{equation*}
where $\widehat{Y}$ represents the set of values to forecast. Here, the posterior predictive distribution is not conditioned on parameter estimates or the inclusion/exclusion of predictors, all of which have been integrated out. Algorithm \ref{algo:forecast} conducts joint quantile prediction, where  forecasts are generated by the Bayesian model averaging approach which provides a coherent mechanism to handle model uncertainty, by averaging the predicted values from all the models generated in the MCMC model training. 
Through Bayesian model averaging, we commit neither to any particular set of predictors which helps avoid an arbitrary selection, nor to point estimates of their coefficients which prevents overfitting.

\section{Model Performance with Simulated Data}
\label{sec:Model_Performance_with_Simulated_Data}
In this section, we demonstrate the model performance with simulated data generated by three-dimensional target time series given in Section \ref{sec:Generated_Data}, in terms of selecting the correct variables and accuracy in parameter estimation in Section \ref{sec:Model_Training_Performance}, and forecast performance of the model with different quantiles and different time series correlations in Section \ref{sec:Forecast_performance}. The AL  likelihood is known to be very restrictive in modeling the underlying error distributions and unlikely to be the true data generating likelihood \citep{yang2016posterior}, so it is primarily used as a working likelihood in the Bayesian quantile regression literature. Even though its posterior consistency for coefficient estimation has been established by \cite{sriram2013posterior} under model misspecification, the validity of using AL likelihood for prediction purpose still remains questionable. Hence, we use error terms simulated by the AL likelihood for coefficient estimation and by the Gaussian distribution for response prediction.

\subsection{Generated Data}
\label{sec:Generated_Data}
The simulated data is generated by the following three-dimensional model (i.e., $m=3$)
\begin{align} 
\label{eq:simulated_model}
\widetilde{y}_t=\widetilde{\mu}_t+\widetilde{\iota}_{t}+B^T\widetilde{x}_t+\widetilde{\epsilon}_t, 
\end{align}
where each time series has its own trend component, seasonal component, and regression component. We will mainly uses the dataset size $n=500$.
The trend component $\widetilde{\mu}_t$ is generated as follows:
\begin{align*}  
\begin{gathered}
\widetilde{\mu}_{t+1}=\begin{bmatrix}
\mu_{t+1}^{(1)}  \\
\mu_{t+1}^{(2)} \\
\mu_{t+1}^{(3)} 
\end{bmatrix}=\begin{bmatrix}
\mu_{t}^{(1)} \\
\mu_{t}^{(2)} \\
\mu_{t}^{(3)} 
\end{bmatrix}+\begin{bmatrix}
\delta_{t}^{(1)}  \\
\delta_{t}^{(2)} \\
\delta_{t}^{(3)} 
\end{bmatrix}+\begin{bmatrix}
u_{t}^{(1)}  \\
u_{t}^{(2)} \\
u_{t}^{(3)}
\end{bmatrix},\quad\quad \begin{bmatrix}
u_{t}^{(1)}  \\
u_{t}^{(2)} \\
u_{t}^{(3)}
\end{bmatrix}\widesim{\text{iid}}  N_3\Bigg(
\begin{bmatrix}
0  \\
0 \\
0
\end{bmatrix},
\begingroup 
\setlength\arraycolsep{1.8pt}
\begin{bmatrix}
1&0 & 0\\
0&1 & 0\\
0 & 0 & 1
\end{bmatrix}
\endgroup\Bigg).
\end{gathered}	
\end{align*} 
where its slope is generated as 
\begin{align*}  
	\begin{gathered}
		\begin{bmatrix}
			\delta_{t}^{(1)}  \\
			\delta_{t}^{(2)} \\
			\delta_{t}^{(3)} 
		\end{bmatrix}=
		\begin{bmatrix}
			0.04+0.6(\delta_{t-1}^{(1)}-0.04)  \\
			0.05+0.3(\delta_{t-1}^{(2)}-0.05) \\
			0.02+0.1(\delta_{t-1}^{(3)}-0.02) 
		\end{bmatrix}+\begin{bmatrix}
		v_{t}^{(1)}  \\
		v_{t}^{(2)} \\
		v_{t}^{(3)}
	\end{bmatrix},\quad\quad \begin{bmatrix}
	v_{t}^{(1)}  \\
	v_{t}^{(2)} \\
	v_{t}^{(3)}
\end{bmatrix}\widesim{\text{iid}}  N_3\Bigg(
\begin{bmatrix}
0  \\
0 \\
0
\end{bmatrix},
\begingroup 
\setlength\arraycolsep{1.8pt}
\begin{bmatrix}
1&0 & 0\\
0&1 & 0\\
0 & 0 & 1
\end{bmatrix}
\endgroup\Bigg).
	\end{gathered}	
\end{align*} 
The seasonal component $\widetilde{\iota}_t$ is generated as follows:
\begin{equation*} \label{eq:sim-seasonal}
	\begin{gathered}
		\tilde{\iota}_{t+1}=\begin{bmatrix}
			\iota_{t+1}^{(1)}  \\
			\iota_{t+1}^{(2)}\\
			\iota_{t+1}^{(3)}
		\end{bmatrix}=\begin{bmatrix}
			-\sum_{k=0}^{100}\iota_{t-k}^{(1)}  \\
			-\sum_{k=0}^{70}\iota_{t-k}^{(2)}  \\
			-\sum_{k=0}^{40}\iota_{t-k}^{(3)}  \\
		\end{bmatrix}+\begin{bmatrix}
			w_{t}^{(1)}  \\
			w_{t}^{(2)}  \\
			w_{t}^{(3)}  
		\end{bmatrix},\quad \begin{bmatrix}
		w_{t}^{(1)}  \\
		w_{t}^{(2)} \\
		w_{t}^{(3)}
	\end{bmatrix}\widesim{\text{iid}}  N_3\Bigg(
	\begin{bmatrix}
		1  \\
		1 \\
		1
	\end{bmatrix},
	\begingroup 
	\setlength\arraycolsep{1.8pt}
	\begin{bmatrix}
		0.5&0 & 0\\
		0&0.5 & 0\\
		0 & 0 & 0.5
	\end{bmatrix}
	\endgroup\Bigg).
	\end{gathered}
\end{equation*}
The regression component $B^T\widetilde{x}_t$ is generated with $8$ explanatory variables, at least one of which has no effect on each target series with zero regression coefficient, as follows:
\begin{align*} 
\begin{gathered}
B=\begin{bmatrix}
2  &  4 &-3.5 &-2  &  0 & 0& -1.6 &   0\\
3 &   0 & 2.5 &-3  &  0 &-1.5 & 0 &   2\\
-2.5  &  0 &-2& -1 &   3 & 2&  0 &   4
\end{bmatrix}^T, \\
\widetilde{x}_t=\begin{bmatrix}
x_{t1} & x_{t2} & x_{t3}&  x_{t4}& x_{t5} & x_{t6}& x_{t7} & x_{t8}
\end{bmatrix}^T,\\
x_{t1}\widesim{\text{iid}} \mathcal {N}(5,2),\ \ \  x_{t2}\widesim{\text{iid}} \operatorname {Pois}(10),\ \ \ x_{t3}\widesim{\text{iid}} \operatorname {Pois}(5),\ \ \  x_{t4}\widesim{\text{iid}} \mathcal {N}(-2,5),\\
x_{t5}\widesim{\text{iid}} \mathcal {N}(-5,2),\ \ \  x_{t6}\widesim{\text{iid}} \operatorname {Pois}(15),\ \ \  x_{t7}\widesim{\text{iid}} \operatorname {Pois}(20),\ \ \ x_{t8}\widesim{\text{iid}} \mathcal {N}(0,10).
\end{gathered}
\end{align*}


To examine the accuracy of variable selection and parameter estimation in Section \ref{sec:Model_Training_Performance}, we use the simulated data generated by the following MAL distributed error term $\widetilde{\epsilon}_t\widesim{\text{iid}} \mathcal{AL}_m(\phi_\epsilon,\Sigma_\epsilon)$ according to equation \eqref{eqn:link} where,  for $\tau=(\tau_1,\tau_2,\tau_3)$,
\begin{equation}
	\label{eqn:error1}
\quad\quad \phi_\epsilon=\Phi \widetilde{\phi}_{\tau} ,\quad\quad  \Sigma_\epsilon=\Phi\Sigma_{\tau}\Phi=\Phi(\Psi_{\tau}\Sigma_{\text{corr}}\Psi_{\tau})\Phi,
\end{equation}

$$
\Sigma_{\text{corr}}=\begin{bmatrix}
1 & \rho & \rho \\
\rho & 1  & \rho \\
\rho & \rho  & 1
\end{bmatrix},
\quad\quad 
\Phi=\begin{bmatrix}
0.7 & 0   & 0 \\
0 & 0.6  & 0 \\
0 & 0  & 0.9
\end{bmatrix},$$
$$
\widetilde{\phi}_{\tau}=
\begin{bmatrix}
\frac{1-2\tau_1}{\tau_1(1-\tau_1)} \\\\
\frac{1-2\tau_2}{\tau_2(1-\tau_2)} \\\\
\frac{1-2\tau_3}{\tau_3(1-\tau_3)}
\end{bmatrix},
\quad\text{and}\quad 
\Psi_{\tau}=\begin{bmatrix}
\sqrt{\frac{2}{\tau_1(1-\tau_1)}} & 0   & 0 \\
0 & \sqrt{\frac{2}{\tau_2(1-\tau_2)}}  & 0 \\
0 & 0   & \sqrt{\frac{2}{\tau_3(1-\tau_3)}}
\end{bmatrix}.
$$
To examine the prediction strength in Section \ref{sec:Forecast_performance}, we use the simulated data generated by the following normal distributed error term 
\begin{equation}
	\label{eqn:error2}
\widetilde{\epsilon}_t\widesim{\text{iid}} N_3\Bigg(
\begin{bmatrix}
	0 \\
	0 \\
	0
\end{bmatrix},
\begingroup 
\setlength\arraycolsep{1.8pt}
\begin{bmatrix}
1 & \rho & \rho \\
\rho & 1  & \rho \\
\rho & \rho  & 1
\end{bmatrix}
\endgroup\Bigg).
\end{equation}

\subsection{Model Training Performance}
\label{sec:Model_Training_Performance}

\begin{figure}
	\subfloat[]{\includegraphics[width = 2.5in, height=1.8in]{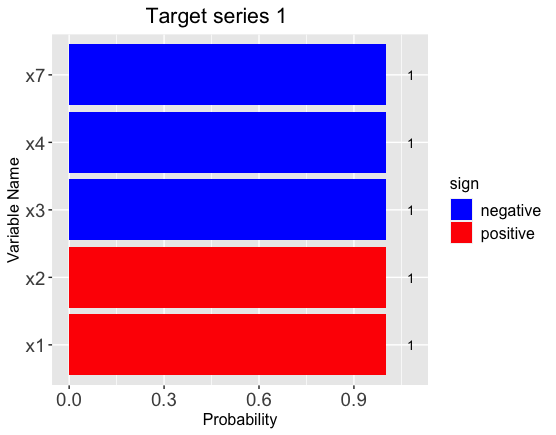}}
		\subfloat[]{\includegraphics[width = 2.5in, height=1.8in]{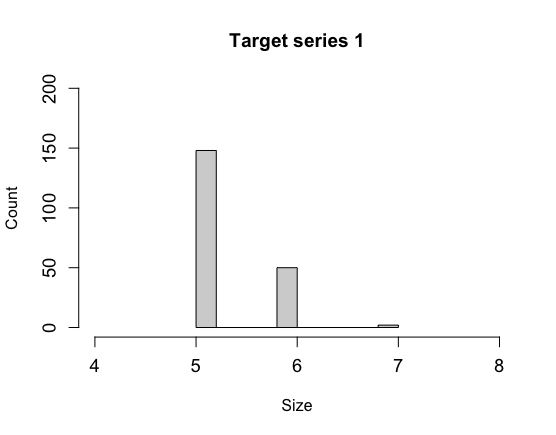}}\\
	\subfloat[]{\includegraphics[width = 2.5in, height=1.8in]{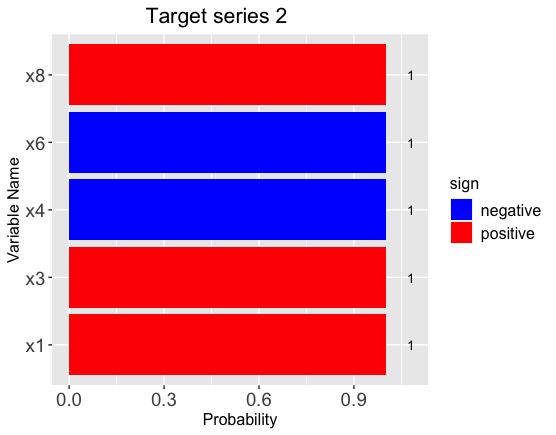}}
		\subfloat[]{\includegraphics[width = 2.5in, height=1.8in]{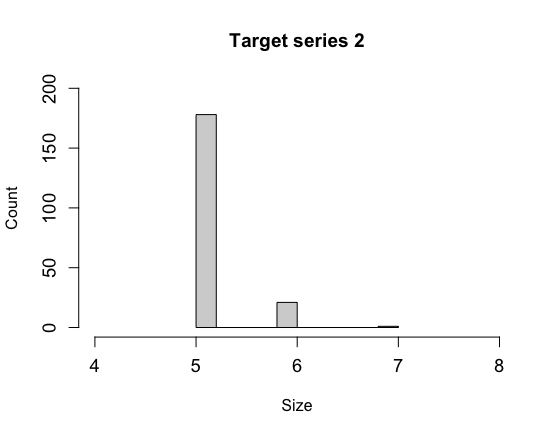}}\\
	\subfloat[]{\includegraphics[width = 2.5in, height=1.8in]{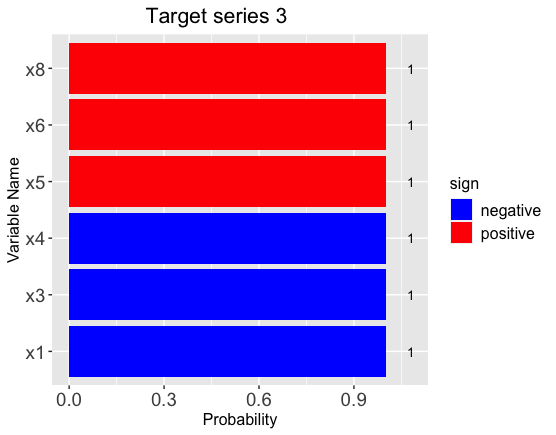}}
	\subfloat[]{\includegraphics[width = 2.5in, height=1.8in]{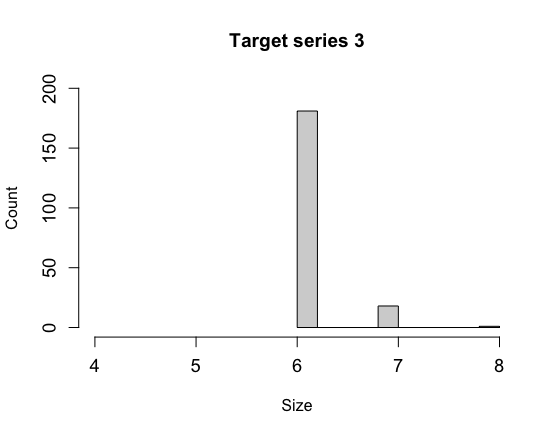}}
	\caption{Feature selection results with simulated data. The $3$-dimensional time series dataset of $500$ observations is generated by Model \eqref{eq:simulated_model} with quantile $\tau=(0.9,0.9,0.9)$ and pairwise correlation $0.7$. The threshold inclusion probability was set as $0.8$. Model training used $400$ MCMC iterations including $200$ discarded as burn-in. The left three plots ((a), (c), and (e)) provide the feature selection results whose X-axis represents the empricial probability (for example, $1$ means that variable was selected $200$ times out of the $200$ MCMC iterations after burn-in), and whose Y-axis represents the variables selected. The right three plots ((b), (d), and (f)) report the count distribution out of these $200$ iterations (Y-axis), where the X-axis stands for the count of selected variables (for example, 5 in (b) corresponds to the 5 variable selected in (a)).}
	\label{fig:Feature_selection}
\end{figure}

In this section, we are going to demonstrate the superior feature selection performances of the QFSTS model with small datasets, using only $400$ MCMC iterations including $200$ discarded as burn-in. The left three plots in Figure \ref{fig:Feature_selection} provide the feature selection results for a dataset of $500$ observations, generated by Model \eqref{eq:simulated_model} with quantile $\tau=(0.9,0.9,0.9)$ and pairwise correlation $0.7$. The threshold inclusion probability was set as $0.8$, i.e., $\geq 80\%$ times a predictor was selected out of the $(400-200)$ MCMC iterations. We can see that the selected features exactly match the model setup, where the value $1$ means a feature was selected all the time out of the $(400-200)$ MCMC iterations. The signs of selected variables also exactly match the model setup, and were marked with red for positive and blue for negative. The right three plots of Figure \ref{fig:Feature_selection} reveal that, for each target series, only in a very small portion out of the $(400-200)$ MCMC iterations, the model selected more variables. Further analyses with different quantiles, correlations, and inclusion probabilities are provided in the Supplementary Material.

\begin{figure}
	\subfloat[]{\includegraphics[width = 2.5in, height=1.8in]{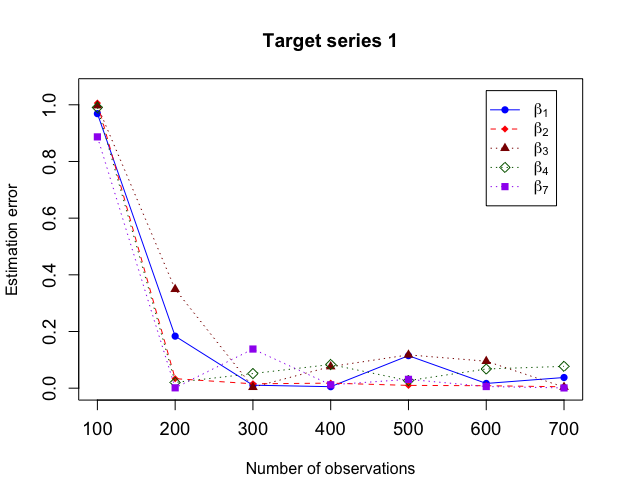}}
	\subfloat[]{\includegraphics[width = 2.5in, height=1.8in]{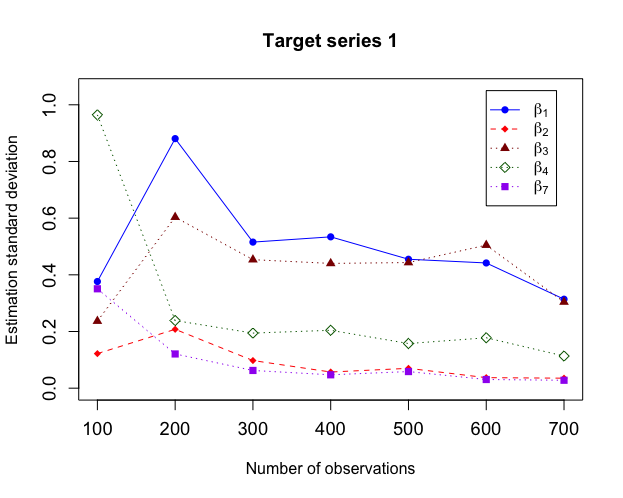}}\\
	\subfloat[]{\includegraphics[width = 2.5in, height=1.8in]{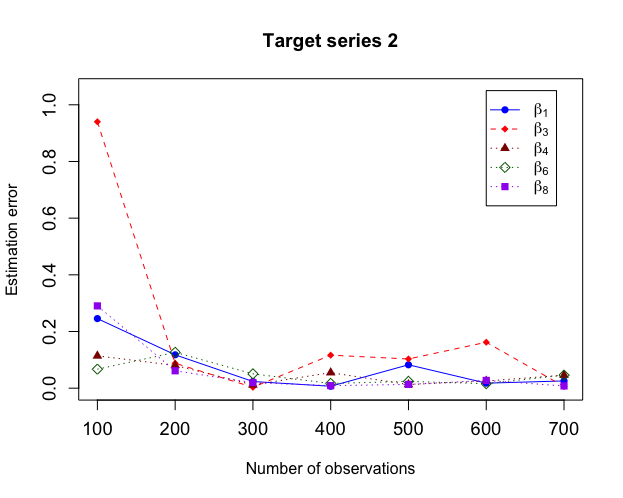}}
	\subfloat[]{\includegraphics[width = 2.5in, height=1.8in]{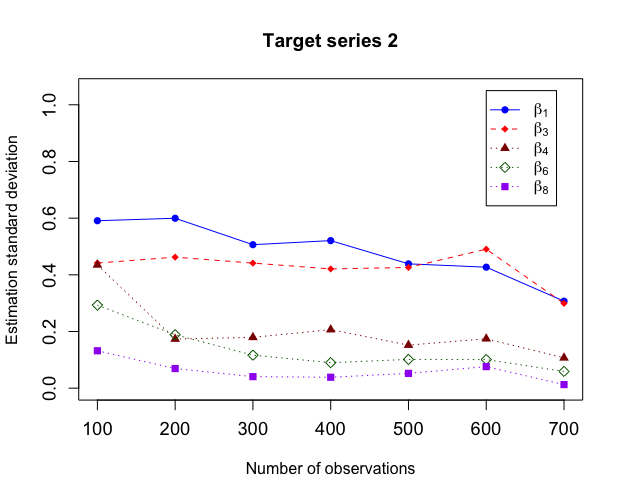}}\\
	\subfloat[]{\includegraphics[width = 2.5in, height=1.8in]{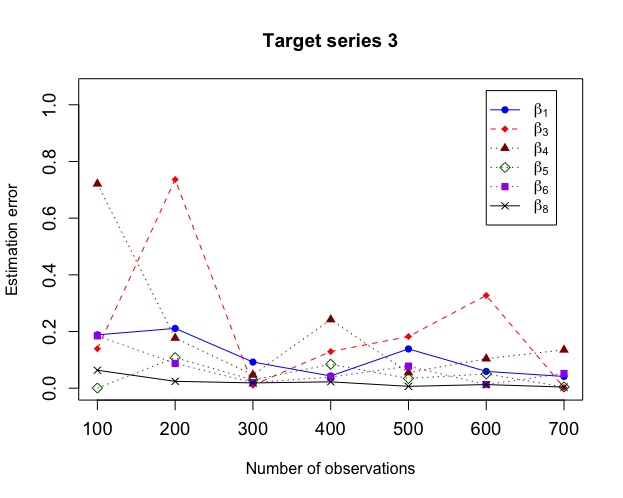}}
	\subfloat[]{\includegraphics[width = 2.5in, height=1.8in]{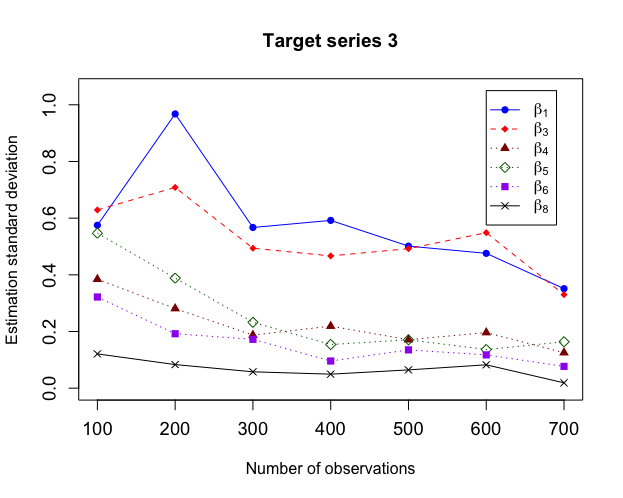}}
	\caption{Estimation errors and standard deviations of regression coefficients. The $3$-dimensional time series datasets were generated by Model \eqref{eq:simulated_model} with quantile $\tau=(0.9,0.9,0.9)$, pairwise correlation $\rho=0.7$, and $7$ dataset sizes ($100$, $200$, $300$, $400$, $500$, $600$, $700$), using only $400$ MCMC iterations including $200$ discarded as burn-in. The left three plots ((a), (c), and (e)) provide the estimation errors of regression coefficients, and the right three plots ((b), (d), and (f)) provide the standard deviations of regression coefficients, for target series $1$, $2$, and $3$, respectively.}
	\label{fig:Estimation_error_sd}
\end{figure}
Figure \ref{fig:Estimation_error_sd} demonstrates the fast convergence and superior parameter estimation performance, with datasets generated by Model \eqref{eq:simulated_model} with quantile $\tau=(0.9,0.9,0.9)$, pairwise correlation $\rho=0.7$, and $7$ dataset sizes ($100$, $200$, $300$, $400$, $500$, $600$, $700$), using only $400$ MCMC iterations including $200$ discarded as burn-in. The left three plots provide the normalized estimation errors calculated as the 
$$|(\text{estimated value} - \text{true value})/\text{true value}|,$$ and the right three plots provide the standard deviations of estimation. 
We can see that both the estimation errors and estimation standard deviations decrease fast as the sample size increases. Similar superior model training performances for different quantiles and correlations are provided in the Supplementary Material.

\subsection{Forecast Performance}
\label{sec:Forecast_performance}

Quantile time series forecasting is the prediction of the distribution of a future
value of a time series. It is much more challenging than the time series mean or median forecast which is already difficult given the additional time information. 
The QFSTS model is a Monte Carlo-based algorithm for quantile prediction. The Monte Carlo samples generate the empirical distribution, whose mean is the quantile prediction.
The most well-known quantile forecast algorithm that is publicly available, is the ``auto.arima" function for univariate time series analysis, in the ``forecast" R package (\cite{Hyndman2008Automatic}).
``auto.arima" automatically fits the best ARIMAX Model, which is the most classical and popular time series model, according to either AIC, AICc, or BIC value. Setting the ``biasadj" option in the ``auto.arima" function to ``FALSE", whose default value is ``TRUE" for mean prediction, gives the quantile prediction. There are only $4$ quantile values possible: $2.5\%$, $10\%$, $90\%$, and $97.5\%$. Therefore, based on these $4$ quantile values, we analyze the QFSTS model's forecast performances.
\begin{figure}
	\subfloat[]{\includegraphics[width = 2.5in, height=1.8in]{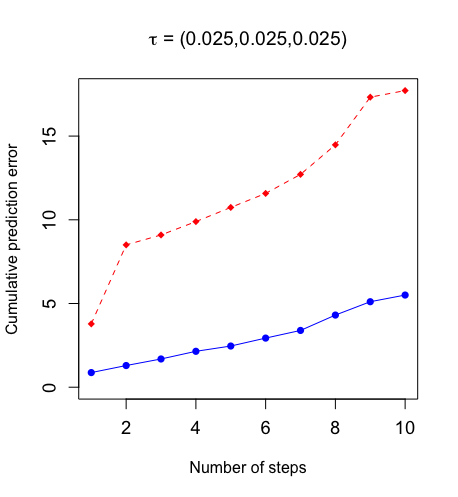}}
	\subfloat[]{\includegraphics[width = 2.5in, height=1.8in]{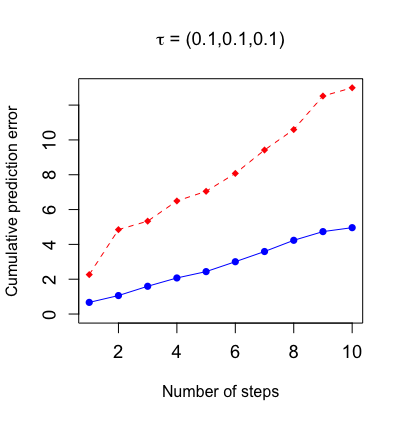}}\\
		\includegraphics[width = 2.4in, height=0.3in]{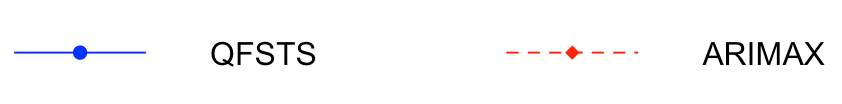}\\
	\subfloat[]{\includegraphics[width = 2.4in, height=1.8in]{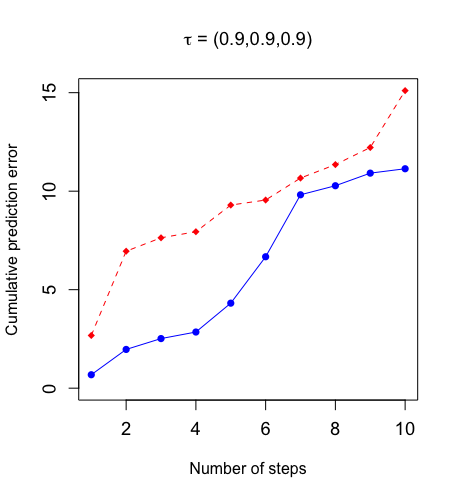}}
	\subfloat[]{\includegraphics[width = 2.5in, height=1.8in]{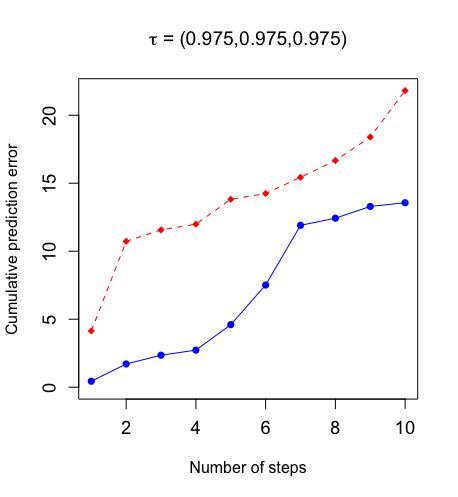}}
	\caption{Cumulative absolute one-step-ahead prediction errors with simulated data for different quantiles. The red colored line is the performance of the ARIMAX model generated by ``auto.arima" function, and the blue colored line is the performance of the QFSTS model.
		Three-dimensional time series datasets of $500$ observations were generated by Model \eqref{eq:simulated_model} with  pairwise correlation $\rho=0.7$. Model training used $400$ MCMC iterations including $200$ discarded as burn-in. Prediction error is measured by the
		quantile loss function given in equation \eqref{eqn:quantile_loss_func}. Plots ((a), (b), (c), and (d)) provide cumulative prediction errors for 
quantiles $\tau=(0.025,0.025,0.025)$, $\tau=(0.1,0.1,0.1)$, $\tau=(0.9,0.9,0.9)$, and $\tau=(0.975,0.975,0.975)$, respectively.}
	\label{fig:forecast_different_tau}
\end{figure}
Figure \ref{fig:forecast_different_tau} reports the forecast performances for three-dimensional target time series datasets of $500$ observations, generated by equation \eqref{eq:simulated_model} with fixed pairwise correlation $\rho=0.7$ but different quantiles: $\tau=(0.025,0.025,0.025)$, $\tau=(0.1,0.1,0.1)$, $\tau=(0.9,0.9,0.9)$, and $\tau=(0.975,0.975,0.975)$. The cumulative prediction error is calculated accumulatively according to the quantile loss function given in \eqref{eqn:quantile_loss_func}, where this standard approach can also be seen in \cite{chen2013bayesian}. We can see that the QFSTS model outperforms ``auto.arima" consistently in the tens steps of one-step ahead forecast.

\section{Conclusion}
\label{sec:conclusion}
In this paper, we have proposed the QFSTS model for joint quantile analysis of correlated time series with dimension $m>1$. 
The correlation matrix of multivariate time series is usually very hard to estimate, while users of QFSTS do not need to evaluate the correlation among series. Hence, running QFSTS for multiple series is simpler than running it for time series one by one, with fewer user efforts involved. The IW prior is commonly used as the prior for the covariance matrix. In the distributional sense, the covariance matrix in the current setting is proportional to the correlation matrix. Hence, given the challenging situation that we allow each time series to have its own time series components (trend and season) and its own pool of predictors, we used the IW prior for simplicity. 
For univariate time series (the variance is just one constant and there is no correlation anymore), the current prior setup is not optimal but its proper design is out of the scope of this paper. 

QFSTS achieves accurate quantile feature selection utilizing the spike-and-slab Bayesian prior setup, which uses the Bernoulli prior distribution for indicator variables. If no specific prior information on the initial inclusion probabilities of particular variables is available, the Bernoulli prior distribution is a common default choice. Variant spike-and-slab modeling can be used instead, such as spike-and-slab being normal distributions (\cite{george1993variable}) or scale mixtures of normals (\cite{ishwaran2005spike}). The QFSTS model uses the standard Metropolis-Hastings algorithm in the MCMC model training while its speed and performance may be further improved with advanced Metropolis-Hastings algorithms (\cite{banterle2019accelerating,sherlock2017pseudo,atchade2005improving,atchade2011towards}). As the first multivariate time series model with joint quantile feature selection, the QFSTS model sheds light on this new research area and outperforms the classical ARIMAX time series model consistently.

Probabilistic forecasts play a fundamental role in addressing the inherent uncertainty in data, supporting decision-making under uncertainty, evaluating and refining forecasting models, and enabling informed decision-making in various domains \citep{gneiting2007probabilistic}. Our methodology is motivated by the asymmetric Laplace working likelihood. It can also be extended to represent functional coefficients using basis representations, allowing for the borrowing of strength from nearby locations and incorporating a global-local shrinkage prior on the basis coefficients to achieve adaptive regularization \citep{liu2020function, Liu2020on}.  Also motivated by a working Laplace likelihood approach, the 
 Bayesian median autoregressive model proposed in \cite{zeng2021bayesian} adopts a parametric model bearing the same structure as autoregressive models by altering the Gaussian error to Laplace, leading to a simple, robust, and interpretable modeling strategy for time series forecasting. Lastly, we acknowledge that the QFSTS model produces point joint quantile estimation, rather than joint interval predictions which is a challenging task that we leave for future research.
 
\bibliographystyle{ba}
\bibliography{sample}

\end{document}